\documentclass{article}

\usepackage[preprint]{neurips_2026}

\usepackage[utf8]{inputenc} % allow utf-8 input
\usepackage[T1]{fontenc}    % use 8-bit T1 fonts
\usepackage{hyperref}       % hyperlinks
\hypersetup{hypertexnames=false} % unique anchors across deferred floats/appendix
\usepackage{url}            % simple URL typesetting
\usepackage{booktabs}       % professional-quality tables
\usepackage{amsfonts}       % blackboard math symbols
\usepackage{nicefrac}       % compact symbols for 1/2, etc.
\usepackage{microtype}      % microtypography
\usepackage{xcolor}      % colors

\usepackage{natbib}
\usepackage{tcolorbox}
\usepackage{CJKutf8}
\usepackage{enumitem}
\usepackage{amsmath,amssymb}
\usepackage{algorithm}
\usepackage{algpseudocode}
\usepackage{graphicx}
\usepackage{cleveref}
\usepackage{placeins} % keep floats within their semantic section
\usepackage{flafter}           % do not place floats before their source location
\usepackage{wrapfig}
\usepackage{multirow}
\usepackage{multicol}
\usepackage{footmisc}

\newcommand{\bottomnote}[1]{%
  \begingroup
  \renewcommand{\thefootnote}{}%
  \footnotetext{\hspace{-1em}#1}%
  \endgroup
}

\title{LiveMem: Maintaining Memory State Continuity in Long-Running LLM Inference}
\author{
  \textbf{Zhichen Liu}$^{1,2}$\thanks{Corresponding authors: \texttt{liuzc2024@mail.sustech.edu.cn}; \texttt{sunruihan2019@163.com}}
  \quad \textbf{Ruihan Sun}$^{1}$\footnotemark[1]
  \quad \textbf{Hengjie Yang}$^{1}$
  \quad \textbf{Zipeng Wu}$^{1,3}$
  \\[0.2em]
  \textbf{Zhaohan Chen}$^{1}$
  \quad \textbf{Xiaofan Zhang}$^{1}$
  \quad \textbf{Yang Xu}$^{2}$
  \\[0.5em]
  \normalfont
  $^{1}$NatureSelect.AI
  \qquad
  $^{2}$Southern University of Science and Technology
  \qquad$^{3}$Xidian University
}

\begin{document}

\maketitle
\begin{abstract}
Long-running assistants and agents consume interaction streams that eventually outgrow the context. Existing context retention, summarization, and retrieval preserve access to selected history, but do not provide a persistent state over the full lifecycle when working context changes. We formulate this missing inference capability as \emph{state continuity under context turnover}: carrying computation forward through a fixed-capacity memory state whose lifetime is independent of the active context. We introduce an intrinsic memory method, \textbf{LiveMem}, which augments a pretrained full-attention LLM with a memory state that preserves the historical information over the whole lifecycle while the main attention path retains a bounded KV window. Context turnover and memory state maintaining, memory-oriented post-training, and state-aware serving jointly make this memory state load bearing after its originating tokens are released. 
Our experiments show that LiveMem achieves leading overall performance among evaluated systems and other intrinsic memory methods. Experiments on LongMemEval show that LiveMem is able to answer the question based on the memory state, even when the supporting evidence has been removed from the current context, and evidence-distance analysis shows that useful information persists beyond the active window. 
LiveMem thus establishes state continuity as a distinct and complementary abstraction for continual LLM inference.
\end{abstract}

\bottomnote{
\includegraphics[height=1.05em]{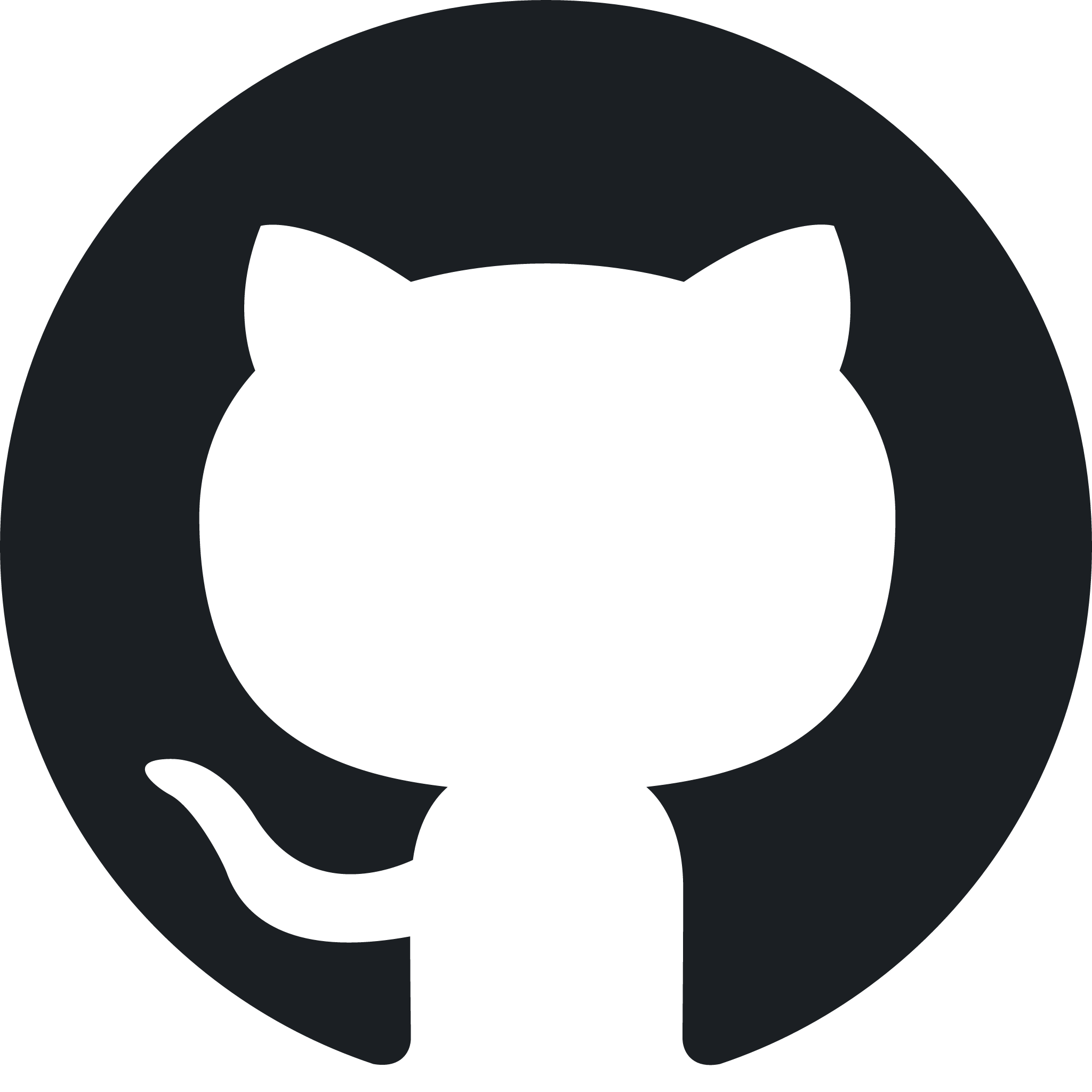} Code Release: \url{https://github.com/cafeii/LiveMem}
}

\section{Introduction}

Language models are evolving from isolated question-answering systems into continuously active assistants and agents over long-term interactions \citep{openclaw, hermes}. Such systems no longer receive a single instruction to complete one task; instead, they continuously interact with the environment and accomplish various tasks in a scenario with no termination signal. As a result, the total history is determined by the lifecycle of the interaction, and has no theoretical ending.
As history grows, the model's working context must inevitably be replaced through some means.
This raises a fundamental question: \emph{how can language models maintain continuity across different working contexts within the same lifecycle?}

\begin{figure}[htbp]
    \centering
    \includegraphics[width=\textwidth]{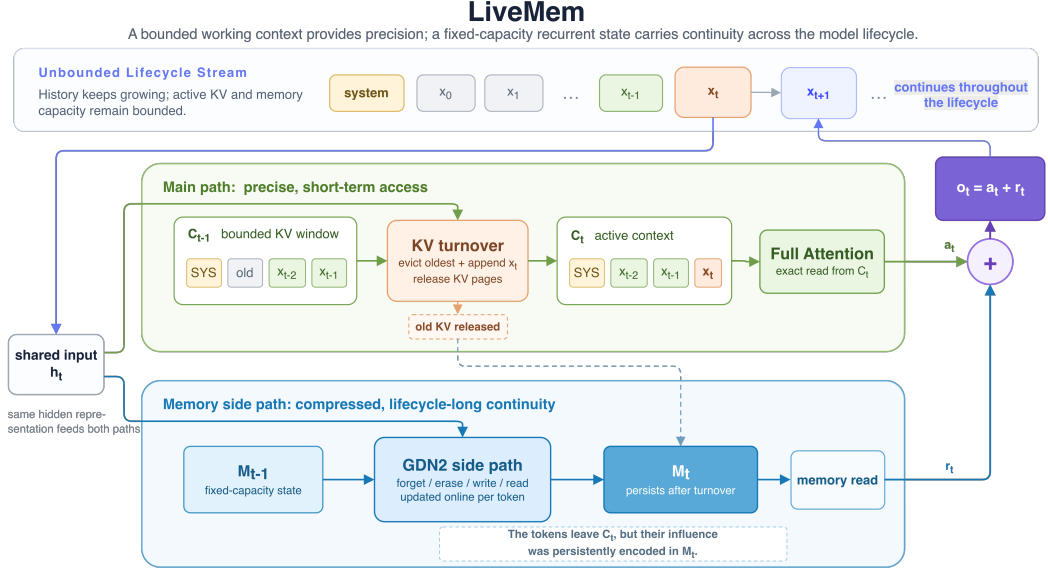}
    \caption{LiveMem enables stateful continual inference under context turnover. A bounded working context provides precise short-term access, while a fixed-capacity recurrent memory state preserves information across the model lifecycle.}
    \label{fig:livemem-main}
    \vspace{-1em}
\end{figure}

Existing widely used RAG-like systems do not truly solve the problem of continuity. Instead, they rely on engineering-based memory management strategies as patches (e.g., high-dimensional summaries) to maintain a certain level of awareness of the entire lifecycle \citep{mem0,memos,amem,hindsight}. These systems provide effective ``historical access'' mechanisms that determine how past content is supplied to the model, we regard them as persistent memory storage like notepads. Since the working context can be viewed as a precise short-term memory mechanism, this distinction reveals a scale missing from existing directions: a continuous memory state throughout the lifecycle—updated online, with fixed capacity, and persisting independently of the working context.

We term this requirement, which exists at the lifecycle scale and is independent of the working context, \emph{state continuity}.
Engineered memory systems are effective at persistent storage and on-demand retrieval, but difficult to provide this seamless continuity.
The latter is precisely what recurrent memory states excel at, which can transmit information from multiple working contexts through model space in a high-dimensional form continuously. Therefore, memory systems that provide ``historical access'' and recurrent memory states that provide ``state continuity'' are not competing definitions of memory; they are orthogonal and potentially complementary in temporal scale and information representation.

Recent research has explored the continuity of memory states at this scale, ranging from neural long-term memory in Titans to the continuous memory perspective in Nested Learning\citep{behrouz2025titans,behrouz2025nested}.
However, merely having a nominal recurrent memory state does not answer this question. If only a recurrent state is used as the model backbone, the model struggles to maintain precision when handling short-term tasks.
Some works introduce a memory state branch on top of the transformer backbone, which provides certain cross-context representation capabilities\citep{yang2026indexmem}, but if key history remains visible to attention at all times, the model may learn to bypass the memory state rather than maintaining continuity through it.
Therefore, establishing a functional continuous memory state requires more than a state tensor labeled as memory: the model must undergo context turnover, learn to effectively write to and read from the memory state, and maintain predictive capability through the memory state even after historical contexts are no longer accessible. Thus, our goal is not merely to add a recurrent mechanism, but to ensure that the memory state genuinely carries effective information. Unlike methods that primarily introduce a memory representation, we explicitly study the lifecycle transition where information leaves attention and must remain usable through a persistent state.

We introduce \textbf{LiveMem}, an attention-LLM augmented with a recurrent memory state maintained by a parallel memory branch using Gated DeltaNet-2 (GDN2) \citep{hatamizadeh2026gdn2} as backbone. The branch continuously read and update a fixed-size memory state, while the original attention path supplies exact access to a bounded working context. KV turnover releases chunks from the attention path when the context reaches its limit, while the information remaining stays in the memory state. We apply memory-oriented post-training task to train the model to write, preserve, and read the memory state. LiveMem is a more complete structure of stateful continual inference: architecture, post-training, context turnover, and serving jointly make the memory state behaviorally effective after its originating context has disappeared.

Our experiments demonstrate the effectiveness of LiveMem. LiveMem achieves leading overall performance across evaluated tasks. Meanwhile, our experiments on LongMemEval show that even when evidence has been completely phased out of the current context window, LiveMem still maintains an accuracy above the baseline; this accuracy varies with the distance from the evidence to the window, but remains measurable. Our experiments indicate that LiveMem is not universally optimal across all tasks—retrieval memory and parametric memory also have their own strengths—but the results sufficiently show that memory states can carry historical information beyond the working context. In summary, our contributions include:
\begin{itemize}[leftmargin=1.3em, itemsep=-0.1em, topsep=-0.1em]
    \item Defining memory state continuity, modeling this continuity with LiveMem architecture, providing an end-to-end design from formulation to implementation.
    \item Proposing memory-oriented post-training tasks to enable memory states to genuinely carry effective information.
    \item Demonstrating the usefulness of LiveMem through comprehensive evaluations and exploring the boundaries of its capabilities.
\end{itemize}

\section{Preliminaries: Inference with Memory State under Context Turnover}
\label{sec:problem}
For continuously running language models, the total token consumption can far exceed their context capacity. Therefore, the issue is not merely how past information is accessed, but rather how to maintain computational continuity when context turnover occurs. This chapter formalizes this problem and defines how latent states can function as memory.

\subsection{Bounded Context and Reconstructive Inference}
\label{sec:reconstructive-inference}

Let $H_t=(x_1,\ldots,x_t)$ denote the interaction history available by time $t$, and let $C_t$ denote the representation of tokens within the context window under the attention mechanism. An attention-only decoder with fixed parameters can be abstracted as
\begin{equation}
    y_t \sim p_\theta(y_t \mid C_t).
    \label{eq:reconstructive-inference}
\end{equation}
Thus, $C_t$ can be viewed as the inference state of the transformer. This state, composed of key-value pairs, grows with the history and, once it exceeds the context window, the information it carries also disappears \citep{kwon2023pagedattention,zhang2023h2o}. Therefore, when $C_t$ leaves the context, historical information can only be reused to influence subsequent predictions by building a system that retrieves $C_t$ or some representation derived from it in some form \citep{lewis2020rag,packer2023memgpt}. This paradigm, in which the model reconstructs its working state by reading explicit representations of relevant past information, is referred to as reconstructive inference.

\subsection{Memory State Continuity}
\label{sec:stateful-inference}

\begin{figure}[tbp]
    \centering
    \includegraphics[width=\textwidth]{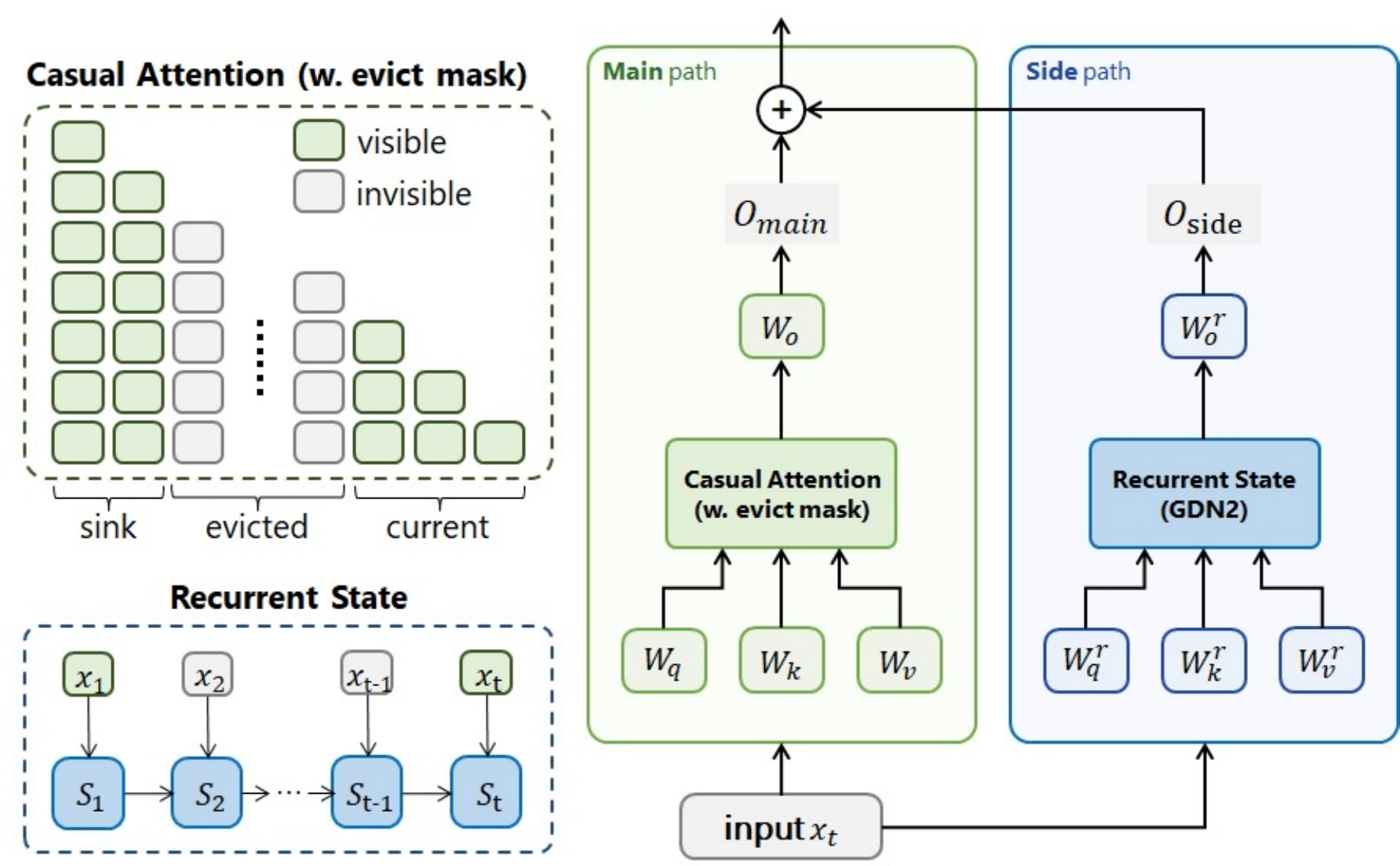}
    \caption{Overview of LiveMem. A full-attention main path maintains a bounded working context, while a recurrent side path updates the recurrent memory state from the same hidden representations. Their outputs are combined at every decoder layer. Context turnover evicts old KV entries from the main path, keeps only the system instruction as attention sink, and the current context in KV cache. The information carried by the evicted context is kept in the recurrent memory state.}
    \label{fig:livemem-overview}
    \vspace{-1em}
\end{figure}

Let $M_t$ be the memory state after the model has processed up to time $t$. We model the update of the memory state as follows:
\begin{align}
M_{t+1} &= U_\phi(M_t, X_{t+1}),
\label{eq:state-update}\\
y_t &\sim p_{\theta,\phi}(y_t \mid C_t, M_t),
\label{eq:stateful-prediction}
\end{align}
where $U_\phi$ is a learned state transition function, and $\phi$ denotes parameters introduced by the memory module. The working context $C_t$ provides fine-grained access to recent information, and $M_t$ carries compressed historical memory information whose size is fixed. We do not require that $M_t$ be able to reconstruct every token exactly. We refer to this setup as continual inference with an accompanying memory state, where inference is viewed as an ever-evolving state transition process that persists throughout the entire lifespan. This mechanism is complementary to external retrieval systems. A standalone memory retrieval system, when the model has no knowledge of the past, can only recall based on the model's understanding of the current query. If the model is equipped with an accompanying memory state, it allows the model to have a rough awareness of the past.

Recurrent memory tokens, compressed attention states, and fixed latent memory pools implement similar memory states \citep{bulatov2022rmt,munkhdalai2024infiniattention,wang2024memoryllm}, but these states rely on independent context-turnover operations. In contrast, we integrate $U_\phi$ into the model itself and train it as part of the model backbone -- namely, the GDN2 side-path architecture that we adopt.

The mere presence of a recurrent module does not prove that it carries memory. We refer to a recurrent state that satisfies these requirements as a living memory state. (i)~it is updated online with incoming inputs, (ii)~it persists after context switches, and (iii)~the critical information contained within it influences the model's subsequent behavior. Therefore, we consider a model to truly carry memory only when it can rely on the memory state to make decisions after the evidence has been completely removed from the context.

\section{LiveMem}
\label{sec:method}

LiveMem adds a recurrent memory path to the attention module of a pretrained decoder transformer, which operates in parallel with the attention computation. These two paths serve different timescales: the attention mechanism provides precise access to the working context, while the recurrent memory path carries memory information across the entire lifetime. In this chapter, we will introduce the overall paradigm and how to train it.

\subsection{Recurrent Memory State and Side Attention}
\label{sec:side-attention}

The memory module of LiveMem is attached as a side branch to each attention layer. It reads and maintains the memory state via Gated DeltaNet-2 (GDN2) \citep{hatamizadeh2026gdn2}, and its output is directly added back to the original attention output. Let $h_t^\ell$ be the normalized hidden representation at layer $\ell$, $S_t^\ell$ be the memory state. The block computes

\begin{align}
    a_t^\ell &= \operatorname{Attn}_\ell
        (h_{\leq t}^\ell; C_t), \\
    (r_t^\ell,S_t^\ell) &= \operatorname{GDN2}_\ell
        (h_t^\ell,S_{t-1}^\ell), \\
    o_t^\ell &= a_t^\ell+r_t^\ell,
    \label{eq:parallel-memory-block}
\end{align}
after which the $o_t^\ell$ returns to the original model. In this way, historical memory information is injected into the model through the side branch of the attention module.

For one recurrent head, let $S_t\in\mathbb{R}^{d_k\times d_v}$ denote its state (omitting layer and head indices). GDN2 produces a log-decay $g_t\leq 0$, an erase gate $b_t\in[0,1]^{d_k}$, and a write gate $w_t\in[0,1]^{d_v}$. Its update can be written as
\begin{align}
    \overline S_t
        &= \operatorname{Diag}\!\left(\exp(g_t)\right)S_{t-1},
        && \text{forget,} \label{eq:gdn2-forget}\\
    \rho_t
        &= w_t\odot v_t
        -\overline S_t^{\top}(b_t\odot k_t),
        && \text{erase and prepare,} \label{eq:gdn2-erase}\\
    S_t
        &= \overline S_t+k_t\rho_t^{\top},
        \qquad r_t=S_t^{\top}q_t,
        && \text{write and read.} \label{eq:gdn2-write}
\end{align}
The decay operation selectively forgets the current memory state along the key dimension, and the erase operation selectively removes old information to reduce interference when writing new information; these behaviors are learned during training. The read and update operations are performed per token to ensure real-time memory utilization and updating, so when context turnover occurs, no additional write operations are performed.

The $q$, $k$ and $v$ parameters of the side branch are initialized from those of the attention backbone, while other parameters are newly initialized. The output projection is initialized with zeros, so that the introduction of the entire side branch is initially equivalent to the original model itself, and the model gradually opens the side branch through learning. See \Cref{sec:memory-parameterization} for details.

\subsection{Live Memory under Context Turnover}
\label{sec:turnover-method}

LiveMem processes the input as an ordered sequence of chunks $X_0,X_1,\ldots$. We retain the system prompt, which structurally serves as an attention sink \citep{xiao2024streamingllm} and semantically preserves the most important system instructions, ensuring that they are always precisely visible to the model. All other chunks processed through a bounded attention window, which is actually a first-in-first-out queue. When a new chunk enters the window, if the total number of chunks in the window or the total number of tokens contained in all chunks exceeds a threshold, old chunks are evicted until it satisfies the threshold. This behavior can be expressed by the following formula:
\begin{equation}
    \operatorname{visible}(i,j)
      = \mathbf{1}[i \le j]\;
        \mathbf{1}\bigl[\text{in\_mem}(\text{chunk}(i), j)\bigr].
    \label{eq:turnover-visibility}
\end{equation}
where $j$ is the position of the current query token, $i$ denotes the position of another token, $c(i)$ represents the chunk containing token $i$, and 
in\_mem(chunk($i$),$j$) indicates whether chunk $i$ is within the window when the query is at position $j$. This behavior ensures that the activation boundary changes only at chunk boundaries, and guarantees that no out-of-bounds situations occur.
\cref{eq:turnover-visibility} applies only to the main attention path, whereas \cref{eq:gdn2-forget,eq:gdn2-erase,eq:gdn2-write} scan over the entire sequence. When an old chunk leaves $C_t$, its KV becomes invisible in the main attention, but its information has already been preserved in the memory state.

We apply this constraint during both training and inference to ensure training-inference consistency. During training, we implement it via dynamic attention masks. During inference, we directly release the evicted KV pages to make them inaccessible. Detailed explanations of the infrastructure are provided in \Cref{sec:infrastructure}

\subsection{Memory-Oriented Post-Training}
\label{sec:post-training}

Adding recurrent states does not guarantee that the model will actually utilize it to carry memory. Therefore, we design memory-oriented post-training tasks. In these tasks, each example consists of a segment of historical memory and one or more corresponding queries. We leverage the context turnover mechanism described in \Cref{sec:turnover-method} to ensure that most of the history is removed from attention during training, thereby forcing the model to learn how to utilize the information in the memory state to produce answers, and also how to better preserve information into the memory state. We train the full memory branch while keeping the backbone parameters frozen. Meanwhile, we also explore training the memory branch with LoRA \citep{hu2022lora}, and examine the effects of partially introducing the backbone parameters into training (\Cref{sec:ablations}). We design two training paradigms: supervised fine-tuning (SFT) and reinforcement learning (RL).

\paragraph{Answer-based Supervised Learning}
% For supervised fine-tuning (SFT), each example concatenates the history,
% query, and reference answer into one causal stream.  Loss is applied only to
% assistant answer positions $\mathcal A$:
In SFT, each example concatenates the history, queries, and reference answers into a single causal stream. The loss is applied only to the answer positions $\mathcal A$:
\begin{equation}
    \mathcal L_{\mathrm{SFT}}
      =-\frac{1}{|\mathcal A|}\sum_{t\in\mathcal A}
        \log p_{\theta,\phi}(x_t\mid C_t,S_t).
    \label{eq:sft-objective}
\end{equation}
We first use long-document question-answering data to establish stable memory-utilization behavior, and help activate the memory side branch. We then train on a broader mixture of data including QA, classification, and multiple-query tasks. Through SFT, the model learns how to use the side branch to carry memory information.

\paragraph{Exploration with Memory}
In RL, we optimize using the GRPO algorithm \citep{shao2024deepseekmath}, incorporating practices from DAPO \citep{yu2025dapo}: clip-higher, token-level loss, dynamic sampling, and removal of KL. Given a prompt $q$ containing memory, we sample $G=8$ responses $o_i$ and compute their advantages with group normalization:
\begin{equation}
    \widehat A_i = \frac{R_i-\mu_R}{\sigma_R},\qquad
    \mu_R=\frac{1}{G}\sum_{j=1}^{G}R_j,\quad
    \sigma_R=\operatorname{Std}_{j}(R_j).
    \label{eq:grpo-advantage}
\end{equation}
For response token $o_{i,t}$, let $\rho_{i,t}=\pi_\theta(o_{i,t}\mid q,o_{i,<t})/ \pi_{\theta_{\mathrm{old}}}(o_{i,t}\mid q,o_{i,<t})$ and $\bar\rho_{i,t}=\operatorname{clip}(\rho_{i,t}, 1-\epsilon_{\mathrm{low}},1+\epsilon_{\mathrm{high}})$. Our maximization objective is
\begin{align}
    J_{\mathrm{RL}}(\theta)
    &= \mathbb E\!\left[
       \frac{1}{\sum_i T_i}\sum_{i=1}^{G}\sum_{t=1}^{T_i}
       \Phi(\rho_{i,t},\widehat A_i)\right],
       \label{eq:dapo-grpo-objective}\\
    \Phi(\rho,A)
    &=
    \begin{cases}
       \min(\rho A,\bar\rho A), & A\geq 0,\\
       \max\!\left(\gamma A,\min(\rho A,\bar\rho A)\right), & A<0,
    \end{cases}
    \nonumber
\end{align}
where $\epsilon_{\mathrm{low}}=0.20$,
$\epsilon_{\mathrm{high}}=0.28$, and $\gamma=3$.
We perform token-level normalization based on the number of tokens in the generated responses. For dynamic sampling, we retain only groups with $\operatorname{Std}_i(a_i)>0$, where $a_i$ is the accuracy of the $i$-th response. Each $q$ may contain one or more questions. We compute accuracy as $a_i=k_i/n_i$, where $n_i$ is the number of questions and $k_i$ is the number of correctly answered questions. When $n_i>1$, we additionally give a bonus of 0.5 if all questions are answered correctly. If there is a formatting issue ($F_i=0$), we apply a penalty of -0.5. Thus, the reward is formulated as:
\begin{equation}
R_i= a_i+0.5 \cdot \mathbf 1[k_i=n_i]-0.5 \cdot \mathbf 1[F_i=0]
\label{eq:rl-reward}
\end{equation}
We first compute accuracy using exact match, and for negative cases, we fall back to an LM judge for equivalence judgment to ensure the unbiasedness of the reward signal as much as possible. Through this process, the model learns to better utilize the side branch for carrying memory.

\section{Experiments}
\label{sec:experiments}

Our experiments primarily address three research questions: 
\begin{enumerate}[leftmargin=1.3em, itemsep=-0.1em, topsep=-0.1em]
    \item How does LiveMem perform compared to other text-based memory and intrinsic memory approaches? (\Cref{sec:main-results})
    \item To what extent can LiveMem retain memories outside the window, and how well can these memories be maintained as the lifecycle evolves? (\Cref{sec:distance-analysis})
    \item How can one effectively train a LiveMem module? (\Cref{sec:ablations})
\end{enumerate}

\begin{table}[tbp]
    \centering
    \caption{Results under bounded-context configurations. The best results on each line are bolded, the second results are underlined.}
    \resizebox{\textwidth}{!}{%
    \begin{tabular}{l|c|cc|cc|cc}
    \toprule
         & Qwen3-4B & RAG & Recurrent & Context2LoRA & $\delta$-Mem & LiveMem-SFT & LiveMem-RL \\
    \midrule
         \multicolumn{8}{c}{Wiki QA} \\
    \midrule
    2Wiki & \underline{0.817} & 0.117 & 0.775 & 0.213 & \textbf{0.824} & 0.768 & 0.784 \\
    2Wiki-MQ & 0.336 & - & 0.204 & 0.009 & 0.232 & \underline{0.551} & \textbf{0.587} \\
    Hotpot & \underline{0.846} & 0.399 & 0.819 & 0.234 & \textbf{0.853} & 0.755 & 0.772 \\
    Hotpot-MQ & 0.329 & - & 0.256 & 0.008 & 0.226 & \underline{0.477} & \textbf{0.487} \\
    MuSiQue & \underline{0.461} & 0.117 & \textbf{0.483} & 0.095 & \textbf{0.483} & 0.369 & 0.383 \\
    MuSiQue-MQ & 0.200 & - & 0.165 & 0.003 & 0.151 & \underline{0.327} & \textbf{0.347} \\
    \midrule
    All & 0.498 & 0.211 & 0.450 & 0.094 & 0.462 & \underline{0.541} & \textbf{0.560} \\
    \midrule
        \multicolumn{8}{c}{Conversation} \\
    \midrule
    LoCoMo & 0.255 & \textbf{0.405} & 0.175 & 0.140 & 0.256 & 0.295 & \underline{0.314} \\
    LME-s & 0.224 & \textbf{0.548} & \underline{0.386} & 0.168 & 0.230 & 0.286 & 0.302 \\
    FactCons & \textbf{0.177} & 0.090 & 0.100 & 0.081 & 0.171 & \underline{0.175} & \underline{0.175} \\
    \midrule
    All & 0.219 & \textbf{0.348} & 0.220 & 0.130 & 0.219 & 0.252 & \underline{0.264} \\
    \midrule
        \multicolumn{8}{c}{TTL} \\
    \midrule
    Banking & 0.770 & 0.740 & 0.000 & 0.860 & 0.770 & \underline{0.920} & \textbf{0.940} \\
    CLINC & 0.870 & 0.660 & 0.000 & 0.830 & 0.890 & \textbf{0.930} & \underline{0.920} \\
    NLU & 0.760 & 0.600 & 0.010 & \textbf{0.860} & 0.770 & \underline{0.820} & \underline{0.820} \\
    TREC-C & 0.730 & 0.720 & 0.340 & \textbf{0.940} & 0.740 & \underline{0.780} & 0.770 \\
    TREC-F & 0.460 & 0.290 & 0.220 & \textbf{0.860} & \underline{0.510} & 0.480 & 0.470 \\
    Redial & 0.141 & 0.069 & 0.081 & 0.081 & \underline{0.146} & 0.131 & \textbf{0.149} \\
    \midrule
    All & 0.622 & 0.513 & 0.109 & \textbf{0.739} & 0.638 & 0.677 & \underline{0.678} \\
    \midrule
        \multicolumn{8}{c}{Long QA} \\
    \midrule
    $\infty$Bench-QA & 0.117 & 0.228 & \textbf{0.316} & 0.083 & 0.114 & 0.234 & \underline{0.251} \\
    EventQA & 0.445 & \textbf{0.612} & 0.447 & 0.269 & 0.433 & 0.463 & \underline{0.506} \\
    NarrQA & 0.313 & 0.233 & 0.282 & 0.150 & 0.326 & \underline{0.332} & \textbf{0.371} \\
    \midrule
    All & 0.282 & \underline{0.358} & 0.348 & 0.167 & 0.291 & 0.343 & \textbf{0.376} \\
    \midrule
        \multicolumn{8}{c}{Overall} \\
    \midrule
      Overall & 0.458	& 0.389	& 0.281	& 0.327	& 0.451 & \underline{0.505} & \textbf{0.519} \\
    \bottomrule
    \end{tabular}
    }
    \label{tab:bounded-context-results}
    \vspace{-1em}
\end{table}

\subsection{Experimental Setup}
\label{sec:experimental-setup}
\paragraph{Tasks and metrics.}
We evaluated our approach on Wiki QA, Conversation QA, Test-Time Learning (TTL), and Long Document QA (Long QA). For the Wiki QA suite, we used 2WikiMultiHopQA, HotpotQA, and MuSiQue as evaluation sets \citep{ho2020twowiki,yang2018hotpotqa,trivedi2022musique}, and additionally packed multiple samples together to form a multi-question (MQ) task. For the Conversation suite, we employed LoCoMo, LongMemEval, and the FactConsolidation dialogue data from MemoryAgentBench (MAB) for evaluation \citep{maharana2024locomo,wu2024longmemeval,hu2026memoryagentbench}. For the TTL suite, we used the six TTL subsets from MAB, namely Banking77, CLINC150, NLU, TREC-coarse/fine, and ReDial \citep{banking77, clinc150, nlu, trec, redial}. For Long QA, we adopted the English QA task from $\infty$Bench \citep{zhang2024infinitebench}, EventQA from MAB, and NarrativeQA \citep{kocisky2018narrativeqa}. We report per-question accuracy, using LLM-Judge to determine whether open-ended answers are consistent with the reference answers, using exact match to judge whether the answers to classification tasks are correct, and using recall@5 for ReDial. Detailed introductions to the evaluation tasks can be found in \cref{sec:additional-experiments}.

\paragraph{Evaluation protocol.}
All systems are compared using Qwen3-4B-Instruct-2507\footnote{\url{https://huggingface.co/Qwen/Qwen3-4B-Instruct-2507} \label{fn:qwen3-4b}} (hereinafter referred to as Qwen3-4B) \citep{yang2025qwen3}. To fully test the performance of each approach under limited context, we restrict the context window to 32k tokens; for shorter datasets (Wiki QA suite, LoCoMo), we further restrict it to 8k tokens. This means that if the compared method does not have a corresponding memory mechanism, the part of the input memory document that exceeds the window limit will be truncated from the head. For all generation, we uniformly adopt the sampling parameters officially recommended by Qwen: temperature 0.7, top-p 0.8, and top-k 20\footref{fn:qwen3-4b}.

\paragraph{Systems.}
We use Qwen3-4B as the direct baseline for comparison, while $\delta$-mem \citep{lei2026deltamem} adds a side-path state to it. We also compare against Context2LoRA, for which we construct the LoRA parameter memory module following the official behavior in our experiments \citet{back2026loramemory}. RAG uses Qwen3-Embedding-0.6B \citep{qwen3embedding} as the retrieval model and retrieves relevant evidence via vector recall for generation. In addition, we design a Recurrent baseline, which behaviorally references MemAgent \citep{yu2025memagent}: it maintains a memory context, updates the memories within it upon each context turnover, outputs the entire memory after each round of processing, and feeds it into the prompt in the next round of memory processing. For LiveMem, we compare the models obtained from the two training steps described in Section 3.3, namely LiveMem-SFT and LiveMem-RL. More implementation details can be found in \Cref{sec:additional-experiments}.

\begin{table}[tbp]
\vspace{-1em}
    \centering
    \small
    \caption{The results of LiveMem performance w./w.o. historical information stored in memory state. ``State'' processes the complete history and stores them in memory state, whereas ``Trunc.'' sees only the truncated suffix and begins from zero state. $\Delta$ is State minus Trunc.}
    \begin{tabular}{l|ccc|ccc}
    \toprule
        & \multicolumn{3}{c|}{LiveMem-SFT}
        & \multicolumn{3}{c}{LiveMem-RL} \\
    Dataset & Trunc. & State & $\Delta$ & Trunc. & State & $\Delta$ \\
    \midrule
    2Wiki-MQ       & 0.512 & 0.551 & +0.039 & 0.545 & 0.587 & +0.042 \\
    LoCoMo         & 0.277 & 0.295 & +0.019 & 0.299 & 0.314 & +0.015 \\
    LME-S          & 0.296 & 0.286 & -0.010 & 0.294 & 0.302 & +0.008 \\
    Banking        & 0.900 & 0.920 & +0.020 & 0.930 & 0.940 & +0.010 \\
    $\infty$Bench-QA & 0.228 & 0.234 & +0.006 & 0.222 & 0.251 & +0.028 \\
    NarrQA        & 0.332 & 0.332 & +0.000 & 0.357 & 0.371 & +0.014 \\
    \bottomrule
    \end{tabular}
    \label{tab:state-control}
\end{table}

\subsection{Main Results}
\label{sec:main-results}
\Cref{tab:bounded-context-results} reports the experimental results under the limited context window setting.
On the Wiki QA suite, Qwen3-4B and $\delta$-mem both achieve strong performance on the single-question tasks, closely followed by LiveMem and Recurrent. However, on the MQ tasks, the performance of these methods drops dramatically, with only LiveMem maintaining its lead. This indicates that existing baselines are good at handling only a single multi-hop logic chain, but struggle when faced with multiple logic chains simultaneously. As a result, LiveMem outperforms the other methods overall on Wiki QA.

We also observe that RAG and Context2LoRA achieve SOTA results on Conversation and TTL suites, respectively. However, they are too specialized on their suites. They show little performance on tasks other than their respective SOTA ones. On Long QA, although RAG and Recurrent each excel on particular tasks, LiveMem maintains an overall lead in this group by achieving balanced and strong scores across all three tasks.

In summary, LiveMem is a well-rounded method that performs consistently across all tasks. The result shows that LiveMem achieves SOTA results on Wiki QA and Long QA, and while it does not rank first on Conversation and TTL, its performance is second only to the SOTA baselines on these tasks. Meanwhile, other baselines each have their own strengths and weaknesses. Once they encounter tasks they are not good at, their performance almost completely collapses. Therefore, overall, LiveMem substantially outperforms the other baselines.

\subsection{Does the Recurrent State Carry Evicted Evidence?}
\label{sec:distance-analysis}

We compare the performance of LiveMem under two settings: one with historical memory incorporated, and the other with historical information directly truncated. For models that include historical memory in their memory state, they can refer to the history to a certain extent when answering questions; for the truncation setting, without historical information, the model can only rely on the information within the current window to answer questions. \Cref{tab:state-control} reports representative experimental results. It can be observed that for LiveMem-SFT, except for a performance drop on LongMemEval, incorporating historical information leads to improvements on all other datasets. For LiveMem-RL, on the reported datasets, performance consistently improves after incorporating historical information. This may be attributed to the fact that the model after RL is better at maintaining and leveraging historical memories.

\begin{figure}[tbp]
    \centering
    \includegraphics[width=\textwidth]{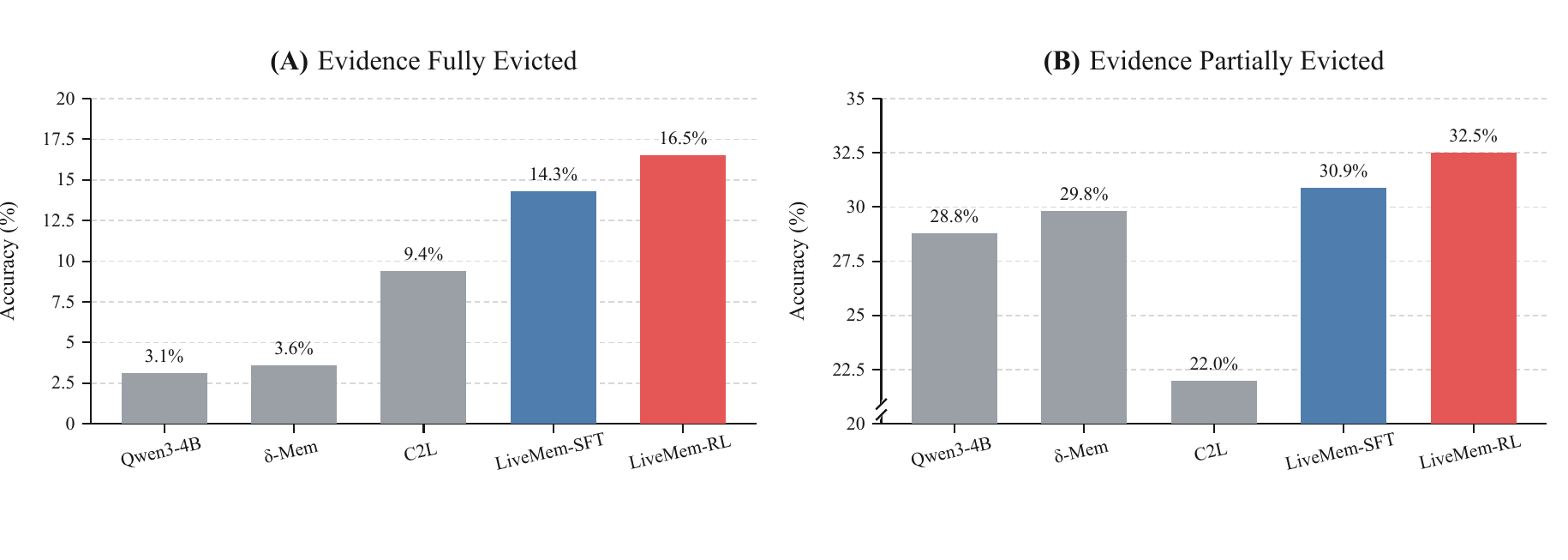}
    \vspace{-3em}
    \caption{LongMemEval-s accuracy grouped by whether the supporting evidence is fully or partially displaced from the active context.  This analysis tests whether the memory state truly benefits performance when the evidence leaves the working context.}
    \label{fig:lme-eviction-results}
    \vspace{-1em}
\end{figure}

\begin{figure}[tbp]
    \centering
    \includegraphics[width=0.6\textwidth]{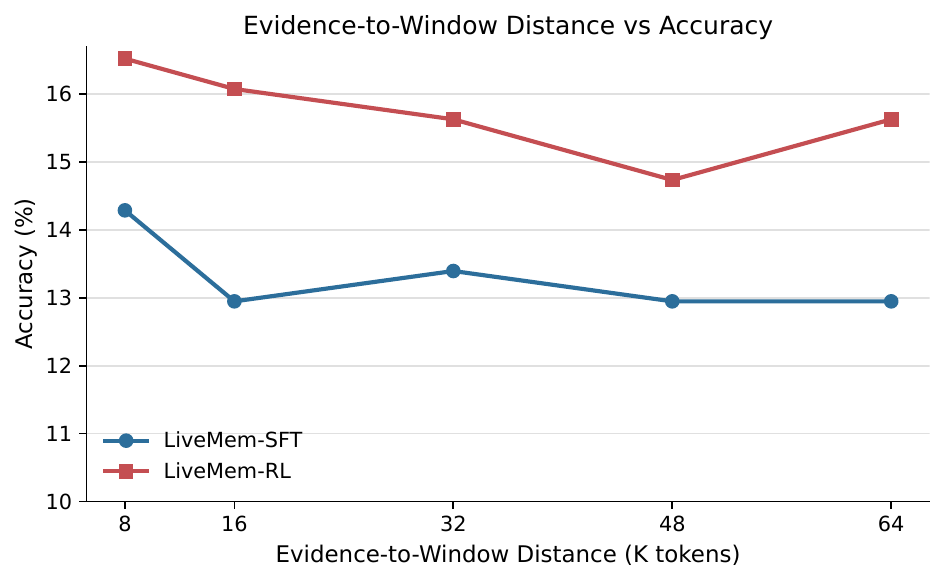}
    \caption{LongMemEval-S accuracy as supporting evidence is placed
    progressively farther from the active-context window.  Accuracy
    remains measurable after multiple turnover intervals rather than
    collapsing at the eviction boundary.}
    \label{fig:evidence-distance}
    \vspace{-1em}
\end{figure}

We further filter the results on LongMemEval and select instances where all key evidence lies outside the window, as well as those where key evidence is partially outside the window, and report the accuracy on these subsets. \Cref{fig:lme-eviction-results} presents the accuracy comparison results. It can be seen that, regardless of whether the evidence is entirely or partially removed, leveraging historical information from LiveMem's memory state can achieve higher accuracy than other methods. Notably, under the full-eviction setting, while Qwen3-4B and $\delta$-mem, which have no access to historical information, achieve extremely low accuracy, LiveMem outperforms them by more than ten percentage points. This demonstrates the necessity of the memory state in carrying historical memories.

We also manually control the distance from evidence to the current window in the LongMemEval subset in \Cref{fig:lme-eviction-results} A to investigate whether the retention duration of evidence in the memory state affects LiveMem's utilization of memories. \Cref{fig:evidence-distance} shows the results as the evicted evidence moves progressively farther away from the active window. The model's accuracy exhibits a slight downward trend as the distance increases, but overall remains at a stable level. This indicates that the memory update and forgetting mechanism based on GDN2 does affect the usability of key evidence in the memory state to some extent, yet it is still capable of keeping such evidence retainable within the memory state.

\subsection{Architecture and Training Ablations}
\label{sec:ablations}

\begin{table}[tbp]
    \centering
    \caption{SFT ablations under bounded active context.  Panel (a) varies
    which parameters are trainable; Panel (b) varies the proportion and order
    of long-context data.}
    \resizebox{\textwidth}{!}{%
    \begin{tabular}{l|ccccccc}
    \toprule
        & 2Wiki-MQ & LoCoMo & LME-S & Banking & TREC-F
        & $\infty$Bench-QA & EventQA \\
    \midrule
    \multicolumn{8}{c}{\textit{(a) Trainable capacity}} \\
    \midrule
    LoRA side                  & 0.479 & \textbf{0.256} & \textbf{0.362} & 0.880 & 0.440 & 0.214 & \textbf{0.490} \\
    Full side           & \textbf{0.551} & \underline{0.234} & \underline{0.286} & \textbf{0.920} & \underline{0.480} & \underline{0.234} & \underline{0.463} \\
    Full side + main-path LoRA & \underline{0.520} & 0.220 & 0.280 & \underline{0.910} & \textbf{0.540} & \textbf{0.236} & 0.412 \\
    \midrule
    \multicolumn{8}{c}{\textit{(b) Training curriculum}} \\
    \midrule
    Direct mixture             & 0.439 & \underline{0.261} & 0.298 & 0.740 & \underline{0.420} & \textbf{0.222} & \underline{0.448} \\
    Rebalanced mixture         & \underline{0.452} & \textbf{0.263} & \underline{0.326} & \underline{0.770} & 0.260 & 0.208 & 0.412 \\
    Long warmup + rebalanced   & \textbf{0.479} & 0.256 & \textbf{0.362} & \textbf{0.880} & \textbf{0.440} & \underline{0.214} & \textbf{0.490} \\
    \bottomrule
    \end{tabular}%
    }
    \label{tab:ablations}
\end{table}

\paragraph{Trainable capacity.}
We compare the performance of LiveMem under different training capacity settings, including side-path LoRA, side-path full-parameter, and side-path full-parameter + main-path LoRA. We present the comparison results of the SFT checkpoints, which are shown in \Cref{tab:ablations} (a). The results indicate that using full-parameter training for the side-path yields more stable performance, likely because the distribution gap between Attention and GDN2 is relatively large, and LoRA—which optimizes within a small subspace—struggles to adapt to the behavior of GDN2. Adding LoRA to the main path does not bring significant improvements; we suspect that the main-path LoRA training may absorb some of the memory that should have been directed to the side-path.

\paragraph{Data Mixture and Training curriculum.}
We also compare the effects of data mixing ratios and curriculum training in \Cref{tab:ablations} (b). Specifically, we compare three settings: (1) using all data directly, (2) using rebalanced data , where we removed the majority of the long-text QA data, which had originally constituted a large proportion of the overall dataset, and (3) using long-text QA data for cold-start to open the side-path, followed by formal training with the rebalanced data. The results show that simply mixing all data together does not perform well, because the long-text data imposes low recall requirements, leaving the model with little incentive to learn how to maintain the memory state. After we reduced the long-text data, the model's performance on recall tasks improved, but its performance on Long QA declined. We attribute that the rebalanced data is too small to sufficiently activate the side-path, which prevents the main path from fully leveraging the information in memory state. Therefore, we first use long-text data for warm-up to activate the side-path, and then use the rebalanced data to train the model to maintain the memory state. This curriculum yields the greatest overall performance improvement.

\section{Conclusion}
What long-context language models truly require goes beyond merely accessing stored historical records; it also entails maintaining coherence of memory when their own working context changes. We formulate this requirement as state continuity under context turnover, thereby distinguishing the model's intrinsic memory state mechanism from external retrieval-based mechanisms. This distinction reveals an inference-time scale that has been overlooked between retrieval systems and attention contexts with limited capacity—the scale over the lifecycle.

LiveMem realizes this mechanism by augmenting a pretrained full-attention model with a parallel GDN2 memory branch to manage the memory state. The memory state is updated online, and when old attention KV is evicted, its information is preserved in the memory state. We propose a standardized scheduling paradigm and training scheme that maintains training-inference consistency under this paradigm while fully training the memory state to bear the burden of memory over the lifecycle.

Across various evaluation tasks, LiveMem demonstrates strong overall performance, and it is particularly effective when multiple questions share the same historical background and the supporting evidence has been entirely moved out of the active window. Ablation studies show that the displaced history still continuously influences predictions through the memory state.

More broadly, retrieval and state continuity address different aspects of long-term reasoning: retrieval determines which past information the model reads, while the recurrent memory state determines how the model maintains information coherence over the lifecycle after reading that information. Therefore, for ultra-long-context scenarios, LiveMem's approach of internally maintaining a memory state to sustain information continuity throughout the entire lifecycle plays an important role in moving toward practical infinite-context inference.

\bibliographystyle{plainnat}
\bibliography{custom}

\clearpage
\appendix

\section{Related Work}
\label{sec:related-work}

\subsection{Textural Memory and RAG-like System}

Retrieval-augmented generation conditions a parametric generator on passages
selected from an external corpus \citep{lewis2020rag}.  Related systems retrieve
model-side representations rather than raw corpus passages.  Memorizing
Transformers performs approximate nearest-neighbor lookup over cached internal
key--value pairs \citep{wu2022memorizing}; LongMem uses a frozen language model
as a memory encoder and trains a residual side network to retrieve and read from
a cached memory bank \citep{wang2023longmem}; and InfLLM stores distant context
blocks in external memory units that are selected for attention without
fine-tuning \citep{xiao2024infllm}.  These methods differ substantially in what
they index and retrieve, but all expand \emph{history access} by selecting an
explicit record or representation to reintroduce into the current computation.

Agent systems additionally manage editable textual records over time.
MemoryBank retrieves, updates, and selectively forgets conversational memories
\citep{zhong2023memorybank}, while MemGPT moves information among context and
external storage tiers \citep{packer2023memgpt}. 
Mem0 \citep{mem0} introduces a scalable long-term memory framework for production-grade agents. By dynamically extracting, consolidating, and retrieving salient facts from historical interactions, it enables agents to maintain consistent user models across multi-turn and cross-session interactions. Its extended version further incorporates graph structures to explicitly capture relationships among memory entries. A-Mem \citep{amem} represents memories as structured notes containing contextual descriptions, keywords, and tags. When new memories are added, it autonomously establishes associations and updates existing memories, thereby forming a continuously evolving memory network. In contrast, MemOS \citep{memos} treats memory as a manageable first-class resource at the system level. Through MemCube, it provides a unified abstraction for textual, activation-based, and parametric memories, while supporting the scheduling, migration, composition, and evolution of different memory forms, thus offering a unified memory infrastructure for continual learning and personalized agents.
These external memory systems rely on heuristic rules and system designs to determine how memories are stored and retrieved, representing a distinct approach from intrinsic memory.

\subsection{Intrinsic Memory}

Past information can also be internalized in model parameters.  Context2LoRA
studies history-specific low-rank adapters as modular knowledge memory
\citep{back2026loramemory}, following the low-rank adaptation mechanism of
\citet{hu2022lora}.  This offers a persistent parametric representation but
requires an optimization procedure to create or update each adapter. 
MLP Memory \citep{wei2026mlpmemory} decouples knowledge storage from the language model decoder by introducing a lightweight, differentiable external MLP module. By imitating a retriever’s behavior over large-scale pretraining corpora, it internalizes non-parametric retrieval patterns into a parametric memory.

Several architectures maintain an explicit latent memory.  MemoryLLM adds a
fixed-size latent memory pool that self-updates from text
\citep{wang2024memoryllm}; M+ augments that design with a co-trained retriever
and longer-term store \citep{wang2025mplus}; and Larimar supports one-shot
knowledge updates through a distributed episodic memory
\citep{das2024larimar}.  Test-Time Training layers make the recurrent hidden
state itself a learned model and update it with a self-supervised objective
\citep{sun2025ttt}.  Titans learns a test-time neural memory, and Nested
Learning generalizes this view to interacting learning processes at multiple
update frequencies \citep{behrouz2025titans,behrouz2025nested}.
$\delta$-Mem augments a frozen full-attention backbone with a compact
associative state whose readout produces low-rank attention corrections
\citep{lei2026deltamem}. In our evaluation protocol, its released evaluation path is nevertheless
context limited: history that exceeds the configured input budget is truncated
before inference, rather than processed through a lifecycle in which old KV
leaves attention while the auxiliary state persists.\footnote{\url{https://github.com/declare-lab/delta-Mem}.}

LiveMem uses the state matrix of an RNN as the memory substrate and maintains the memory state online during inference. It can therefore be viewed as either a parametric method or a form of latent memory. We do not distinguish between these two perspectives here and refer to it simply as intrinsic memory.

\subsection{Recurrent and Hybrid Sequence Models}

Segment recurrence has a long history in Transformer language models.
Transformer-XL reuses hidden states from preceding segments as extended
context \citep{dai2019transformerxl}, whereas Recurrent Memory Transformer
passes learned memory tokens between segments and trains the model to control
their contents \citep{bulatov2022rmt}.  Modern subquadratic models provide
fixed-size recurrent computation through selective state-space models such as
Mamba \citep{gu2023mamba}.  Hybrid architectures combine recurrent and
softmax-attention mechanisms: Infini-attention places local masked attention
and compressive linear memory in one block
\citep{munkhdalai2024infiniattention}, while Jamba interleaves Transformer and
Mamba layers \citep{lieber2024jamba}.

Linear attention supplies another direct connection between recurrence and
memory.  It can be interpreted as a fast-weight programmer with a finite
associative matrix, motivating delta-rule updates that overwrite an existing
key-value association \citep{schlag2021fastweight}.  DeltaNet provides a
parallel training algorithm for this update and studies hybrids with
softmax-attention layers \citep{yang2024deltanet}; Gated DeltaNet combines the
delta rule with a decay gate for rapid forgetting
\citep{yang2025gateddeltanet}; and Kimi Delta Attention adopts finer-grained
gating within a layerwise hybrid \citep{kimi2025linear}.  GDN2 further separates
erase and write controls \citep{hatamizadeh2026gdn2}.  LiveMem uses GDN2 as a
parallel \emph{memory branch} in every decoder layer rather than replacing the
pretrained attention backbone or pretraining a recurrent-attention hybrid from
scratch.  Its focus is also post-training: the recurrent capacity is exposed to
turnover so that its memory state becomes behaviorally useful after source
tokens leave attention.

\subsection{Bounded Attention and KV Turnover}

KV-cache methods bound active attention without necessarily introducing a
carried latent state.  StreamingLLM retains recent tokens together with initial
attention sinks \citep{xiao2024streamingllm}; H$_2$O balances recent entries
with attention heavy hitters \citep{zhang2023h2o}; and SnapKV uses a prompt
observation window to select head-specific KV positions
\citep{li2024snapkv}.  InfLLM moves distant blocks to an external context store
for relevance-based lookup \citep{xiao2024infllm}. 

IndexMem is principally a learned KV-cache eviction method: an importance indexer retains selected KV entries exactly, while evicted entries update a state whose residual readout is trained to compensate for attention removed by compression \citep{yang2026indexmem}. IndexMem employs an indexer to perform query-aware token selection. Similar to the query-conditioned mechanism in MemAgent \citep{yu2025memagent}, it uses the given question to determine which tokens are important. However, applying such a mechanism to general-purpose memory introduces additional uncertainty, because future queries are not yet known when information must be committed to memory. IndexMem primarily focuses on query-aware KV compression rather than maintaining a query-independent memory state throughout a lifecycle. Therefore, it addresses a different problem from state continuity.

LiveMem instead targets this setting by maintaining a query-independent memory state during context turnover: its memory state explicitly carries critical information and remains functional during context turnover.

\section{Reference Results: Under 256K Context Limit}
\label{sec:full-context-results}

We additionally evaluate a context-limited protocol in which up to 256K tokens
remain materialized in the attention context and no pages are released during
the request. Inputs that exceed the model budget lose the oldest memory
tokens while preserving the recent suffix, query, and generation reserve.  This
protocol mirrors the operational behavior of the released $\delta$-Mem
code, which fills a configured context budget and keeps the recent
tail on overflow.

\begin{table}[htbp]
    \centering
    \caption{Performance with a 256K materialized-context cap.  Overflow is
    tail-preserving: the oldest memory tokens are removed before inference.
    This is a context-limited reference rather than a state-continuity test.}
    \resizebox{\textwidth}{!}{%
    \begin{tabular}{l|c|cc|cc|cc}
    \toprule
         & Qwen3-4B & RAG & Recurrent & Context2LoRA & $\delta$-Mem & LiveMem-SFT & LiveMem-RL \\
    \midrule
         \multicolumn{8}{c}{Wiki QA} \\
    \midrule
    2Wiki & 0.683 & 0.117 & \underline{0.775} & 0.218 & 0.637 & 0.771 & \textbf{0.787} \\
    2Wiki-MQ & 0.511 & - & 0.204 & 0.009 & 0.187 & \underline{0.625} & \textbf{0.653} \\
    Hotpot & 0.632 & 0.399 & \textbf{0.819} & 0.237 & 0.590 & 0.756 & \underline{0.772} \\
    Hotpot-MQ & 0.361 & - & 0.256 & 0.025 & 0.144 & \underline{0.706} & \textbf{0.718} \\
    MuSiQue & \underline{0.400} & 0.117 & \textbf{0.483} & 0.096 & 0.365 & 0.362 & 0.379 \\
    MuSiQue-MQ & 0.305 & - & 0.165 & 0.002 & 0.201 & \underline{0.381} & \textbf{0.417} \\
    \midrule
        \multicolumn{8}{c}{Conversation} \\
    \midrule
    LoCoMo & 0.575 & 0.405 & 0.175 & 0.210 & \underline{0.585} & 0.554 & \textbf{0.599} \\
    LME-s & 0.384 & \textbf{0.548} & 0.386 & 0.156 & 0.360 & 0.422 & \underline{0.438} \\
    FactCons & \textbf{0.175} & 0.090 & 0.100 & 0.096 & 0.165 & 0.168 & \underline{0.170} \\
    \midrule
        \multicolumn{8}{c}{TTL} \\
    \midrule
    Banking & 0.830 & 0.740 & 0.000 & 0.910 & 0.860 & \textbf{0.950} & \underline{0.940} \\
    CLINC & 0.860 & 0.660 & 0.000 & \underline{0.920} & 0.850 & \textbf{0.950} & \textbf{0.950} \\
    NLU & 0.800 & 0.600 & 0.010 & \underline{0.820} & 0.790 & \underline{0.820} & \textbf{0.830} \\
    TREC-C & 0.680 & 0.720 & 0.340 & \textbf{0.940} & 0.690 & \underline{0.800} & 0.660 \\
    TREC-F & 0.460 & 0.290 & 0.220 & \textbf{0.870} & 0.460 & \underline{0.570} & 0.510 \\
    Redial & \underline{0.169} & 0.069 & 0.081 & 0.142 & \textbf{0.172} & 0.093 & \underline{0.169} \\
    \midrule
        \multicolumn{8}{c}{Long QA} \\
    \midrule
    $\infty$Bench-QA & 0.291 & 0.228 & 0.316 & 0.131 & 0.302 & \underline{0.407} & \textbf{0.419} \\
    EventQA & \underline{0.624} & 0.612 & 0.447 & 0.252 & 0.608 & 0.515 & \textbf{0.654} \\
    NarrQA & 0.410 & 0.233 & 0.282 & 0.181 & \underline{0.411} & 0.383 & \textbf{0.412} \\
    \bottomrule
    \end{tabular}
    }
    \label{tab:full-context-results}
\end{table}

As demonstrated in \Cref{tab:full-context-results}, increasing the context limit does not fundamentally alter the overall trend. The conclusions are similar to those in \Cref{sec:main-results}: RAG and Context2LoRA continue to perform well on the tasks to which they are best suited, but remain highly task-specific. The key difference is that LiveMem achieves comprehensive improvements over the baselines on LoCoMo and Long QA tasks, indicating that a larger context limit is particularly beneficial to LiveMem. Overall, LiveMem continues to maintain the leading position with strong and well-rounded performance across tasks.

% \Cref{tab:full-context-results} reports the results under the 256K-cap
% reference protocol.  The larger materialized context changes which mechanism is most effective but
% does not erase the overall task dependence seen in the bounded-context table.
% LiveMem-RL remains strongest on all three long-QA datasets and all three
% multi-question Wiki QA tasks, while retrieval remains strongest on
% LongMemEval-S and Context2LoRA on TREC.  Direct context and $\delta$-Mem are
% also competitive when the required evidence remains inside the 256K suffix.
% These results should not be read as evidence that a 256K prompt provides
% persistent memory: once the cap is exceeded, its oldest content is discarded
% rather than carried forward.  The bounded-context and eviction-conditioned
% analyses in the main paper address that distinct question.

\section{Infrastructure and State Lifecycle Details}
\label{sec:infrastructure}

\subsection{GDN2 Parameterization and State Geometry}
\label{sec:memory-parameterization}

The implementation uses a short causal convolution of width four before the
GDN2 query, key, and value projections.  For completeness, the single-head
state update in \cref{eq:gdn2-forget,eq:gdn2-erase,eq:gdn2-write} is equivalent
to
\begin{equation}
    S_t =
    \left(I-k_t(b_t\odot k_t)^\top\right)
    \operatorname{Diag}(\exp(g_t))S_{t-1}
    +k_t(w_t\odot v_t)^\top.
    \label{eq:gdn2-full-update}
\end{equation}
Queries and keys are $\ell_2$-normalized inside the recurrent kernel.  The
readout $S_t^\top q_t$ passes through a gated RMS normalization and an output
projection before it is added to the softmax-attention output.

The Qwen3-4B backbone has 36 layers, 32 query heads, 8 KV heads, and head
dimension 128.  The memory path uses 32 query/key/value heads of dimension 128
in every layer.  Backbone KV projections are repeated across the corresponding
query-head groups for initialization and can diverge during post-training.
Consequently, each layer and request has a recurrent tensor of shape
$32\times128\times128$ in addition to the short-convolution state. The state using FP32.

The query, key, and value projections of the memory branch are initialized from
the matching backbone projections.  The decay, erase, write, output-gate, and
short-convolution parameters are initialized independently, and the final
memory projection is initialized to zero.

\subsection{Chunk-Turnover Training with FlexAttention}

Given chunk spans $[u_i,v_i)$, the canonical policy appends each non-sink
chunk to a live queue and removes the oldest entries while either the chunk
count or token budget is exceeded.  The newest chunk is never removed at its
arrival step.  The policy returns an eviction step for every chunk, which is
expanded into the per-token $c(i)$ and $e(i)$ fields used by
\cref{eq:turnover-visibility}.  SFT examples use dataset-defined semantic or
fixed-token chunks; serving and RL recomputation use deterministic fixed-stride
chunks. Both instantiate the same oldest-first state transition.

During SFT, FlexAttention \citep{dong2024flexattention} combines the turnover
predicate with causality and,
for packed data, a sequence-identity predicate.  The GDN2 chunk kernel receives
cumulative sequence boundaries and starts a zero state at each packed-example
boundary.  Thus a 64K-token packed row can be evaluated in one differentiable
forward pass without allowing either attention or recurrent state to cross
between constituent examples.  Only answer tokens contribute to
\cref{eq:sft-objective}.

RL rollouts use a length-dependent fixed-stride policy.  The trainer derives
the same per-token chunk and eviction fields from the rollout tokens when it
recomputes log probabilities, so the policy update sees the attention
visibility used to generate the response.  Chunk boundaries are multiples of
the 32-token serving page size, which makes token-level mask changes and page
release agree exactly.

Let $L_p$ denote the length of the fully templated prompt and let $M$ be the
generation allowance in a rule-table row.  The rollout server and training
recomputation select the first row of \Cref{tab:turnover-rule-table} satisfying
$L_p+M$ in the indicated half-open interval.  The system-instruction prefix is
rounded up to a 32-token page boundary and retained as a sink; all subsequent
chunks follow the selected stride, including during generation.  For
determinism, even a partially filled frontier chunk is charged at its full
chunk size when enforcing the live-token budget.

\begin{table}[htbp]
    \centering
    \caption{Fixed-stride turnover parameters used for RL rollout serving and
    policy recomputation.  All quantities are tokens.}
    \small
    \begin{tabular}{c|ccc}
    \toprule
    Range of $L_p+M$ & Chunk size & Live-token budget & $M$ \\
    \midrule
    $[1\mathrm{K},2\mathrm{K})$   & 64  & 512   & 256 \\
    $[2\mathrm{K},8\mathrm{K})$   & 256 & 1,024 & 512 \\
    $[8\mathrm{K},16\mathrm{K})$  & 512 & 2,048 & 1,024 \\
    $[16\mathrm{K},32\mathrm{K})$ & 512 & 4,096 & 2,048 \\
    $[32\mathrm{K},64\mathrm{K})$ & 512 & 8,192 & 4,096 \\
    \bottomrule
    \end{tabular}
    \label{tab:turnover-rule-table}
\end{table}

Benchmark inference uses the same oldest-first transition and page-alignment
rules, while overriding the chunk size to 1,024 and the live-token budget to
8K or 32K to realize the fixed evaluation conditions in
\Cref{sec:experimental-setup}.  Samples outside the RL table are filtered
during training; serving clamps an out-of-range request to the nearest row
rather than changing the lifecycle semantics.

\subsection{Serving}

The serving implementation follows paged KV management
\citep{kwon2023pagedattention} and maintains two pieces of per-request state.  The
ordinary attention manager holds KV pages for the instruction sink and recent
live chunks.  A separate paged slot holds the convolution state and GDN2 matrix
state for every memory-augmented layer.  At each turnover boundary, complete
middle KV pages are returned to the allocator and omitted from the attention
read set; the memory slot is unaffected.  Prefill slices are clamped to the
next chunk boundary so a scheduler batch cannot apply a future turnover state
to earlier tokens in the same slice.

For a prefill or decode step, the engine gathers the request's recurrent state,
runs the same GDN2 update as training, and scatters the final state back to its
slot.  Requests without an initial slot read a zero state.  When generation
finishes, the scheduler removes the request's turnover record and the state
manager releases its recurrent slot.  This explicit lifecycle implements
continuity within a stream while preventing accidental state leakage between
requests.  Learned decay and erasure in \cref{eq:gdn2-full-update} operate
within that lifecycle; whole-state reset is a separate systems operation.

\subsection{Post-Training Recipe}

SFT uses fixed 64K-token packs in two stages.  The first stage runs for 100
steps on long-form instruction and document data using LongAlign, LongAlpaca, LongMIT, and Long-Data-Collection \citep{bai2024longalign, chen2024longlora, chen2025longmit, togetherai2023longdatacollections}, with a 50-step learning-rate warmup followed by a constant rate.  The second runs for 500
steps on a mixture of MuSiQue, 2WikiMultihopQA, NarrativeQA, Qasper,
long-form instruction data, AG News, and DBpedia
\citep{dasigi2021qasper,zhang2015charcnn}.  We use an answer-token
cross-entropy objective, AdamW with weight decay 0.01, gradient clipping at
1.0, and a peak learning rate of $5\times10^{-5}$.  In the second stage, the
rate remains constant for 50 steps and then follows a cosine schedule.  LoRA
dropout is zero.

SFT is run in BF16 on two nodes with eight A800-80GB GPUs each.  Each device
processes one 64K pack per step, giving a global batch of 16 packs, or
approximately one million tokens per optimizer step.  We use gradient
checkpointing, distributed data parallelism, and ZeRO-1 optimizer-state
sharding.  The GDN2 recurrent state remains FP32.

The RL stage uses groups of eight responses and the reward described in
\Cref{sec:post-training}.  We use group-relative advantages, no KL penalty,
lower and upper clipping ratios of 0.20 and 0.28, token-mean loss aggregation,
an actor learning rate of $10^{-5}$, and dynamic sampling that discards groups
with zero accuracy variance.  Training uses FSDP2 actors, asynchronous vLLM
rollouts, two eight-GPU trainer nodes, one eight-GPU rollout node, and a
separate two-GPU tensor-parallel judge service. Per-request generation limits are coupled to the
turnover bucket so generated tokens cannot silently exceed the context policy
used for their prompt.

\section{Additional Experimental Details}
\label{sec:additional-experiments}

\subsection{Evaluation Dataset Construction}

2WikiMultiHopQA, HotpotQA, and MuSiQue are knowledge-based question-answering datasets constructed from Wikipedia-derived documents \citep{wiki18}, and they are all multi-hop QA datasets. These datasets evaluate the performance of memory mechanisms in multi-hop knowledge QA scenarios.
Each instance in HotpotQA and 2WikiMultiHopQA consists of 10 documents, which include both relevant and irrelevant documents, with the evidence required to answer the question distributed across the relevant documents. MuSiQue is similar, except that it consists of 20 documents per instance. For the MQ setting, each instance is formed by packing together 5 randomly sampled instances from the corresponding dataset (10 for 2WikiMultiHopQA), meaning that the model must answer all questions simultaneously based solely on a memory summary it has condensed from the packed data.
For each of 2WikiMultiHopQA, HotpotQA, and MuSiQue, we evaluate a fixed seed-0 sample of 2,000 single-question instances and 400 multi-question instances. For the MQ setting, all questions are answered at once after the memory is processed.

LoCoMo, LongMemEval, and FactConsolidation are all dialogue-based QA datasets that evaluate the capability of memory mechanisms in realistic dialogue scenarios.
We use all 500 questions in LongMemEval-S, all questions associated with the ten released LoCoMo conversations, and all questions in MAB FactConsolidation.

The TTL suite is based on existing text-label pairs and requires labeling the input text-label pairs. It demands that the memory mechanism generalize patterns and distill experience from past text-label pairs, rather than simply memorizing the text verbatim. We directly use the MAB's TTL episodes, and follow its evaluation settings.

The LongQA suite focuses on completing question-answering tasks over long input texts, where the information required to answer the questions is scattered throughout the lengthy text. It requires the memory mechanism to possess the ability to gather such distributed information and answer questions when faced with extremely long texts. NarrativeQA is sampled by document: we select 120 test documents with seed 0 and evaluate every retained question for those documents. 

Open-ended QA and conversation answers are evaluated by a Qwen3.6-35B-A3B judge prompted with the question, reference answers, and prediction and asked for a binary correctness decision.  The judge uses greedy decoding with thinking disabled.  EventQA and the five TTL classification sources use normalized exact match.  ReDial uses recall@5 after resolving the released entity identifiers to movie names.

\subsection{Context Accounting and Generation}

Every method receives the same system instruction, task template, and answer
format.  When an input exceeds a direct-context cap, the instruction and query
are retained and the memory portion is truncated from the oldest end.  The
256K condition additionally reserves the task-specific generation allowance
and 1,024 tokens for chat-template overhead before allocating the remaining
budget to memory.  The bounded direct-context controls expose the corresponding
recent 8K or 32K suffix; LiveMem instead scans up to 256K input tokens while
maintaining only the stated live KV budget.

Except for separately identified official-protocol checks, decoding uses
temperature 0.7, top-$p$ 0.8, and top-$k$ 20.  Single-question requests permit
up to 1,024 generated tokens and multi-question requests up to 8,192.  Prompts
that contain multiple questions are never shortened by dropping questions;
the memory prefix is the only truncation target.

\subsection{Baseline Implementation Details}

For the evaluation of all approaches, we construct inputs in the format of [system prompt, user prompt (memory documents, question), assistant answer].
For Qwen3-4B, when the context exceeds the window limit, we retain the system prompt and perform head truncation only on the memory documents.
For $\delta$-Mem, we adopt its TSW setting. Although its paper contains claims that are inconsistent with its code behavior, we choose to strictly follow its code behavior, i.e., directly discarding the portion that exceeds the context window limit.
%rather than additionally implementing a context-turnover state-passing mechanism for it.
For Context2LoRA, for each input memory instance, it pre-constructs multiple QA pairs and then uses these QA pairs to train a LoRA module (rank=4, alpha=8, lr=$5\times 10^{-4}$). During evaluation, it leverages the memories stored in the LoRA parameters to answer questions. We apply a separate LoRA module for each dataset. For TTL, since it inherently consists of text-label pairs and is not suitable for constructing QA pairs, we choose to follow Context2LoRA's original assumption about memory, and directly fine-tune LoRA using these text-label pairs. This may be the reason why Context2LoRA performs particularly well on TTL.
For the RAG method, we use Qwen3-Embedding-0.6B as the vector model and build a vector retrieval environment over the memory documents for retrieval and recall. For datasets with fewer than 20 memory documents, we use top-1; for others, we use top-3. If the input does not have a strict partitioning into memory documents (e.g., Long QA), we split it into chunks of 512 tokens each. Since RAG's top-k fundamentally cannot meet the evidence requirements of the MQ task, we omit RAG's performance on MQ.
For Recurrent, we draw on MemAgent's setting of maintaining a memory context through recurrent cycles. We set the upper limit of the memory context to 8k tokens, so the maximum length of memory text it can process at one time is 16k tokens. The model is required to extract important information from the memory documents and, after processing is complete, answer questions based on the memory context.

\section{Limitations}
\label{sec:limitations}

\subsection{A Lossy State Is Not an Exact Archive}
\label{sec:niah-limitations}

The task variants in this section follow the RULER long-context evaluation
suite \citep{hsieh2024ruler}.

\begin{table}[htbp]
    \centering
    \caption{Needle-in-a-haystack accuracy with full and bounded active
    contexts.  The bounded-context result shows that the current LiveMem model
    does not reliably recover arbitrary fine-grained needles from its recurrent
    state.}
    \begin{tabular}{l|c|c|c|c}
    \toprule
        & S-NIAH-2 & S-NIAH-3 & MK-NIAH-1 & MQ-NIAH \\
    \midrule
    \multicolumn{5}{c}{\textit{256k}} \\
    \midrule
    Qwen3-4B & 1.00 & 1.00 & 0.97 & 0.97 \\
    $\delta$-Mem & 1.00 & 1.00 & 0.97 & 0.97 \\
    LiveMem & 1.00 & 0.86 & 0.99 & 0.97 \\
    \midrule
    \multicolumn{5}{c}{\textit{32k}} \\
    \midrule
    Qwen3-4B & 0.19 & 0.18 & 0.21 & 0.23 \\
    $\delta$-Mem & 0.19 & 0.18 & 0.21 & 0.23 \\
    LiveMem & 0.19 & 0.18 & 0.21 & 0.23 \\
    \bottomrule
    \end{tabular}
    \label{tab:niah-limitations}
\end{table}

The results are shown in \cref{tab:niah-limitations}.  With the 256K
materialized context, where the needles remain directly addressable through
attention, all systems are near ceiling on most variants; LiveMem adds no
systematic benefit. Under the 32K bounded context, all
three systems obtain the same low scores.  The current LiveMem memory state
therefore does not reliably reconstruct arbitrary token-level needles after
their KV entries have been released.

This is consistent with the intended role of the memory state, but it remains
a real limitation.  NIAH asks for lossless archival recall of sparse strings
whose later relevance is not predictable from their content.  LiveMem instead
compresses an indefinitely growing stream into a fixed-size, lossy memory
state and trains that state to preserve features useful for subsequent
behavior.  The evidence-conditioned LongMemEval analyses in
\cref{fig:lme-eviction-results,fig:evidence-distance} show that this state can
carry task-relevant influence across turnover; they do not imply that it stores
every past token verbatim.

The distinction from query-aware KV retention is important here.  IndexMem's
long-prefill indexer can score the prompt after a terminal query is present and
keep the query-relevant needle exactly in attention \citep{yang2026indexmem}.
That is a natural advantage on NIAH, but it does not require a latent state to
carry information whose future relevance was unknown at eviction time.  In a
continuing stream, no fixed KV budget can retain every token that might become
a future needle.  Applications requiring exact archival recall should
therefore combine LiveMem with retrieval or an external store; these mechanisms
provide addressable history access while LiveMem provides state continuity.

\subsection{Finite Positional Horizon}

LiveMem's recurrent memory state and active KV budget are independent of the
number of previously processed tokens, but the current system is not literally
unbounded.  Its Qwen3 backbone uses rotary position encodings \citep{rope} with a configured
maximum position of 262,144 tokens \citep{yang2025qwen3}.  KV eviction preserves
the original absolute position indices of retained tokens, so continuing past
the trained and configured range requires positional extrapolation whose
quality is not guaranteed.  The experiments therefore process at most 256K
input tokens even though the memory state itself could continue updating.

Position-free attention offers a plausible route around this architectural
ceiling.  Kimi Linear applies NoPE to its full-attention layers and delegates
positional and recency information to recurrent KDA layers, demonstrating the
design at 48B total parameters and a 1M-token context
\citep{kimi2025linear}.  Kimi K3 adopts a KDA-based hybrid architecture into a publicly deployed 2.8T-parameter model with a 1M-token
context \citep{kimi2026k3}.  This provides deployment-scale evidence that the
direction is practical.  Adapting
LiveMem to a NoPE or otherwise length-generalizing backbone remains as future work.

% The NeurIPS checklist should follow all appendix material before submission:
% \input{checklist}

\end{document}